\documentclass[lettersize,journal]{IEEEtran}
\usepackage{amsmath,amsfonts}
\usepackage{algorithmic}
\usepackage{array}
\usepackage[caption=false,font=normalsize,labelfont=sf,textfont=sf]{subfig}
\usepackage{textcomp}
\usepackage{stfloats}
\usepackage{url}
\usepackage{verbatim}
\usepackage{graphicx}
\usepackage{booktabs}
\usepackage{multirow}
\usepackage{cite}
\usepackage{amssymb}
\usepackage[colorlinks=true,allcolors=blue]{hyperref}

\def\BibTeX{{\rm B\kern-.05em{\sc i\kern-.025em b}\kern-.08em
    T\kern-.1667em\lower.7ex\hbox{E}\kern-.125emX}}
\usepackage{balance}
\begin{document}
\title{PCFlow: Physics-Conditioned Flow Matching for GPR B-Scan Image Synthesis}

\author{Zhijie~Shen,
        Chenchen~Fu,
        Xuanhao~Chang,
        Hongtao~Bai,
        and~Lili~He%
\thanks{This study was supported by the National Key R\&D Program of China under Grant No. 2022YFF06069003. Corresponding author: Lili He.}%
\thanks{The source code is available at \url{https://github.com/GeometryFu/PCFlow}.}%
\thanks{Z.~Shen, C.~Fu, and X.~Chang are with the College of Computer Science and Technology, Jilin University, Changchun 130012, China.
E-mail: shenzj2123@mails.jlu.edu.cn; fucc2123@mails.jlu.edu.cn; changxh2123@mails.jlu.edu.cn.}%
\thanks{H.~Bai and L.~He are with the College of Computer Science and Technology, Jilin University, Changchun 130012, China, and also with Symbol Computation and Knowledge Engineering of the Ministry of Education, Jilin University, Changchun 130012, China.
E-mail: baiht@jlu.edu.cn; helili@jlu.edu.cn.}%
}


\maketitle

\begin{abstract}
Ground-penetrating radar (GPR) B-scan image synthesis is important for data augmentation, algorithm validation, and simulation acceleration, yet generating radargrams with both visual realism and physical consistency remains challenging. Existing learning-based generative models often emphasize visual appearance but provide limited control over response geometry. In this paper, we propose PCFlow, a physics-conditioned flow matching framework for fast GPR B-scan image synthesis. The core of PCFlow is a Maxwell-informed dense physical condition field constructed from the parameterized physical model used for electromagnetic simulation, including material properties, target geometry, propagation cues, and response-domain priors. This condition field provides an interpretable interface between physical scene parameters and radar response geometry, and guides conditional flow matching in the VAE latent space toward physically feasible generation paths. We evaluate PCFlow on a gprMax-based buried-pipeline dataset with both in-distribution and out-of-distribution test cases. Experimental results show that PCFlow generates images with more accurate response geometry and high visual fidelity, demonstrating its effectiveness for controllable and physically faithful radar image synthesis.
\end{abstract}

\begin{IEEEkeywords}
Ground-penetrating radar, B-scan image synthesis, flow matching, physics-conditioned generation, data augmentation.
\end{IEEEkeywords}

\section{Introduction}
\IEEEPARstart{G}{round}-penetrating radar (GPR) is widely used for non-destructive subsurface sensing, including buried utility mapping, pavement assessment, geological investigation, and underground object detection~\cite{tong2020advances}. A GPR B-scan records time-domain electromagnetic responses along a survey line and provides an image-like representation of subsurface structures. However, acquiring large-scale and well-annotated GPR data is costly and site-dependent, and the collected responses are strongly affected by sensor configurations, environmental conditions, and material variability. Full-wave electromagnetic simulation, such as the finite-difference time-domain (FDTD) method~\cite{gurel2000three,warren2016gprmax}, provides physically reliable radar responses, but large-scale parameter sweeps over materials, geometries, and acquisition settings are computationally expensive.  In practical simulation pipelines, generating one B-scan may take minutes, and can slow down significantly to hours for deeper or more complex scenes. Therefore, fast and physically faithful GPR B-scan image synthesis is important for data augmentation, algorithm validation, simulation acceleration, and learning-based subsurface interpretation.

Learning-based generative models have recently been explored for GPR data augmentation and radar image synthesis, including generative adversarial networks, diffusion models, and conditional image generation frameworks~\cite{fazeel2021gpr,yue2021generation,xiong2023gpr,wang2023gpr,bazrafshan2025synthetic,chen2025elevating}. These methods can improve generation efficiency to seconds-level and produce visually plausible radargrams. Nevertheless, GPR B-scan image synthesis is not only a visual generation problem. For subsurface interpretation, the generated radargram must preserve response geometry determined by target location, burial depth, medium properties, and acquisition configuration. A B-scan with realistic texture may still be physically invalid if the target-induced hyperbolic response appears at an incorrect lateral position, depth, or aperture shape~\cite{lei2019automatic}. Existing learning-based methods often lack such physical controllability. Unconditional models mainly capture the overall radargram distribution and provide limited control over the underlying physical scene. Existing conditional GPR B-scan generation methods are commonly formulated as simulated-to-real image translation, where a simulated B-scan is required as the input~\cite{isola2017image,xu2024fm,pang2026forward}. This leaves an important gap in \textbf{fast, physically consistent GPR B-scan image synthesis from parameterized scene descriptions rather than precomputed radargrams.}


\begin{figure}[!t]
    \centering
    \includegraphics[width=\linewidth]{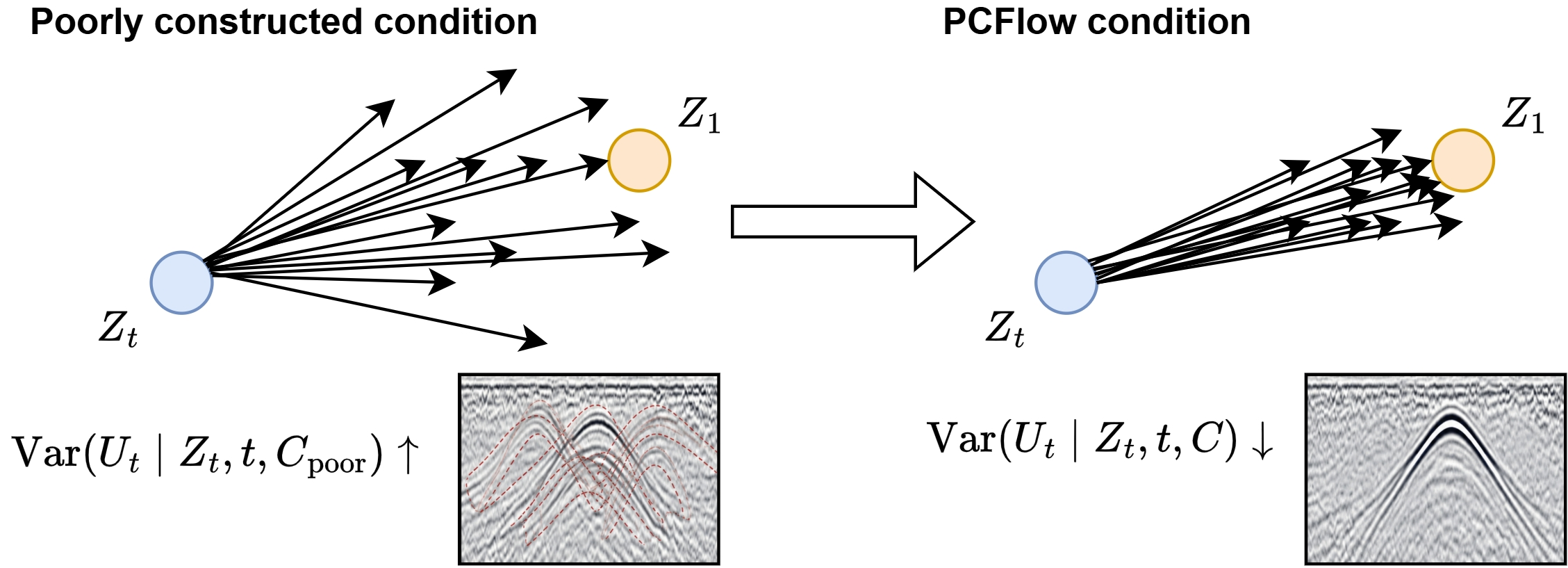}
    \caption{Intuition of variance reduction by the PCFlow condition field. Weak conditioning leads to dispersed velocity targets from the same latent state $\mathbf z_t$, whereas the proposed dense physical condition concentrates the feasible target distribution, reduces $\operatorname{Var}(\mathbf u_t\mid \mathbf z_t,t,\mathbf C)$, and guides the flow toward a physically consistent GPR response.}
    \label{fig:variance_reduction}
\end{figure}

The key challenge is to construct a conditioning representation that connects parameterized physical scenes with radargram-domain responses. In GPR forward modeling, material properties and target geometry determine wave velocity, attenuation, reflection strength, and travel-time curves~\cite{warren2016gprmax,lei2019automatic}. However, these physical factors are usually described in the scene domain, whereas the B-scan is organized by antenna position and propagation time. A suitable condition should therefore be both physically interpretable and radargram-aligned, so that the generator can follow physically feasible response trajectories instead of relying only on appearance-level image statistics (see Fig.~\ref{fig:variance_reduction}).

To address this challenge, we propose PCFlow, a physics-conditioned flow matching framework for fast GPR B-scan image synthesis. Instead of using class labels, text prompts, or existing radar images as the primary condition, PCFlow constructs a dense physical condition field from the parameterized model used for electromagnetic simulation. The condition field combines scene-domain physical cues, such as material properties and target geometry, with response-domain priors, such as approximate travel-time curves, attenuation patterns, and hyperbola anchors. This deterministic encoding strategy provides an interpretable interface between physical scene parameters and radar response geometry. Based on this condition field, PCFlow formulates GPR B-scan image synthesis as conditional probability transport in the VAE latent space, where a flow matching model learns to transport Gaussian latent noise to the latent representation of the corresponding B-scan~\cite{rombach2022high,lipman2023flow}. We further introduce Physics-SPADE modulation and classifier-free guidance to strengthen dense physical conditioning and improve condition-specific generation, especially under distribution shifts~\cite{park2019semantic,ho2021classifierfree}.

We focus on buried pipeline scenarios and construct a physically diverse GPR B-scan dataset based on gprMax~\cite{warren2016gprmax}. The dataset contains approximately 800 pairs of physical conditions and radar responses, covering different soil electromagnetic properties, pipe materials, pipe fillings, target sizes, burial depths, and high-conductivity out-of-distribution cases. Experiments on both in-distribution and out-of-distribution test sets show that PCFlow achieves more accurate response geometry than GAN-based, diffusion-based, ControlNet-style, and text-guided generative baselines, while maintaining strong image fidelity in terms of PSNR, SSIM, and LPIPS~\cite{wang2004image,zhang2018unreasonable}.

The main contributions of this work are threefold:
(1) We address the problem of physically consistent GPR B-scan generation from parameterized physical scene descriptions, and propose PCFlow, a physics-conditioned flow matching framework for controllable and physically faithful radargram synthesis.
(2) We design a deterministic Maxwell-informed condition encoder that converts simulation parameters into dense radargram-aligned fields integrating material, geometry, propagation, and response-domain priors. We further introduce Physics-SPADE grouped modulation to improve condition injection and out-of-distribution (OOD) response-geometry robustness.
(3) We construct a gprMax-based buried-pipeline benchmark and evaluate PCFlow against representative GAN, diffusion, controllable generation, and text-guided baselines. The results show that PCFlow improves both physical response consistency and image fidelity, while highlighting that image-quality metrics alone are insufficient for evaluating physics-conditioned GPR B-scan image synthesis.

\section{Related Work}
\subsection{GPR B-Scan Image Synthesis and Data Augmentation}
GPR B-scan generation has traditionally relied on electromagnetic forward modeling, which can produce physically reliable radar responses but becomes costly when large parameter sweeps over materials, target geometries, burial depths, and acquisition settings are required~\cite{gurel2000three,warren2016gprmax}. To improve data generation efficiency, learning-based GPR B-scan image synthesis and augmentation methods have been increasingly explored~\cite{tong2020advances}. Existing studies commonly use GAN-based, image-to-image, or diffusion-based models to generate or enhance radargrams~\cite{fazeel2021gpr,xiong2023gpr,yue2021generation,wang2023gpr,huang2025sau,hao2025improved,xu2024fm,bazrafshan2025synthetic,chen2025elevating,pang2026forward}. However, most of these methods are either unconditional or only weakly conditioned by random noise, image-domain inputs, or masks. As a result, they can improve visual diversity or realism, but provide limited explicit control over physically important response geometry, such as the lateral position, burial depth, aperture, and attenuation of the target-induced hyperbola. Since these geometric cues are essential for GPR interpretation~\cite{lei2019automatic}, GPR B-scan image synthesis requires not only a realistic radargram appearance but also a controllable interface between physical scene parameters and B-scan responses. PCFlow addresses this issue by introducing a deterministic Maxwell-informed condition encoding strategy that converts material properties, target geometry, propagation cues, and response-domain priors into dense radargram-aligned conditions.

\subsection{Conditional Generative Models and Physical Control}
Most existing learning-based GPR B-scan generators are built on adversarial or diffusion models. GAN-based methods can synthesize radar-like images efficiently, but their training may be unstable, and their outputs are often controlled only indirectly through latent noise or image translation inputs~\cite{isola2017image}. Diffusion models provide more stable training and better sample diversity, and have recently been applied to GPR B-scan generation, restoration, and augmentation~\cite{ho2020denoising,bazrafshan2025synthetic,chen2025elevating,pang2026forward}. Nevertheless, standard diffusion sampling usually requires many denoising steps, and physical consistency still depends heavily on how conditions are represented and injected. In this work, we adopt conditional flow matching, which learns a continuous probability transport from a simple prior to the data distribution and enables efficient generation through deterministic velocity-field integration~\cite{lipman2023flow,liu2023flow}. More importantly, PCFlow couples this framework with a Maxwell-informed dense condition field, Physics-SPADE modulation~\cite{park2019semantic}, and classifier-free guided sampling~\cite{ho2021classifierfree}. The dense condition field and Physics-SPADE inject scene-domain physical quantities and response-domain hyperbola priors into the learned velocity field, while CFG amplifies condition-specific transport during inference. This design allows PCFlow to generate B-scan images that are both visually realistic and physically aligned with the given simulation parameters.

\section{Methodology}

\subsection{Problem Formulation and Overview}
\label{sec:problem}
Let $\mathcal S$ denote a physical simulation condition parsed from a gprMax configuration file. It contains material fields and geometric metadata,
\begin{equation}
    \mathcal S=\{\epsilon_r,\sigma,\mathcal G\},
\end{equation}
where $\epsilon_r\in\mathbb R^{H\times W}$ is the relative permittivity field, $\sigma\in\mathbb R^{H\times W}$ is the conductivity field, and $\mathcal G$ contains target geometry, antenna configuration, simulation domain, and temporal window information. Given $\mathcal S$, our goal is to synthesize the corresponding GPR B-scan image $I_{\mathrm B}$ by modeling $p_\theta(I_{\mathrm B}\mid\mathcal S)$.

PCFlow first constructs a radargram-aligned physical condition field
\begin{equation}
\label{eq:condition_field}
    \mathbf C=\Phi(\mathcal S),
    \qquad
    \mathbf C\in\mathbb R^{K\times H\times W},
\end{equation}
where $\Phi$ is a deterministic Maxwell-informed condition encoder, and $K$ denotes the number of task-specific condition fields. The purpose of $\Phi$ is to bridge the domain gap between the parameterized scene and the radar response: a gprMax configuration describes materials, geometry, and acquisition settings in the physical scene domain, whereas a B-scan is organized by antenna position and wave travel time. Therefore, PCFlow constructs $\mathbf C$ as a radargram-aligned dense condition field that contains both scene-domain physical cues and response-domain priors. The condition field is then downsampled to the VAE latent resolution as $\mathbf C_{\mathrm{lat}}=D_{\downarrow}(\mathbf C)$ and used to guide conditional flow matching in latent space~\cite{lipman2023flow,podell2024sdxl}.

\begin{figure*}[!t]
    \centering
    \includegraphics[width=\textwidth]{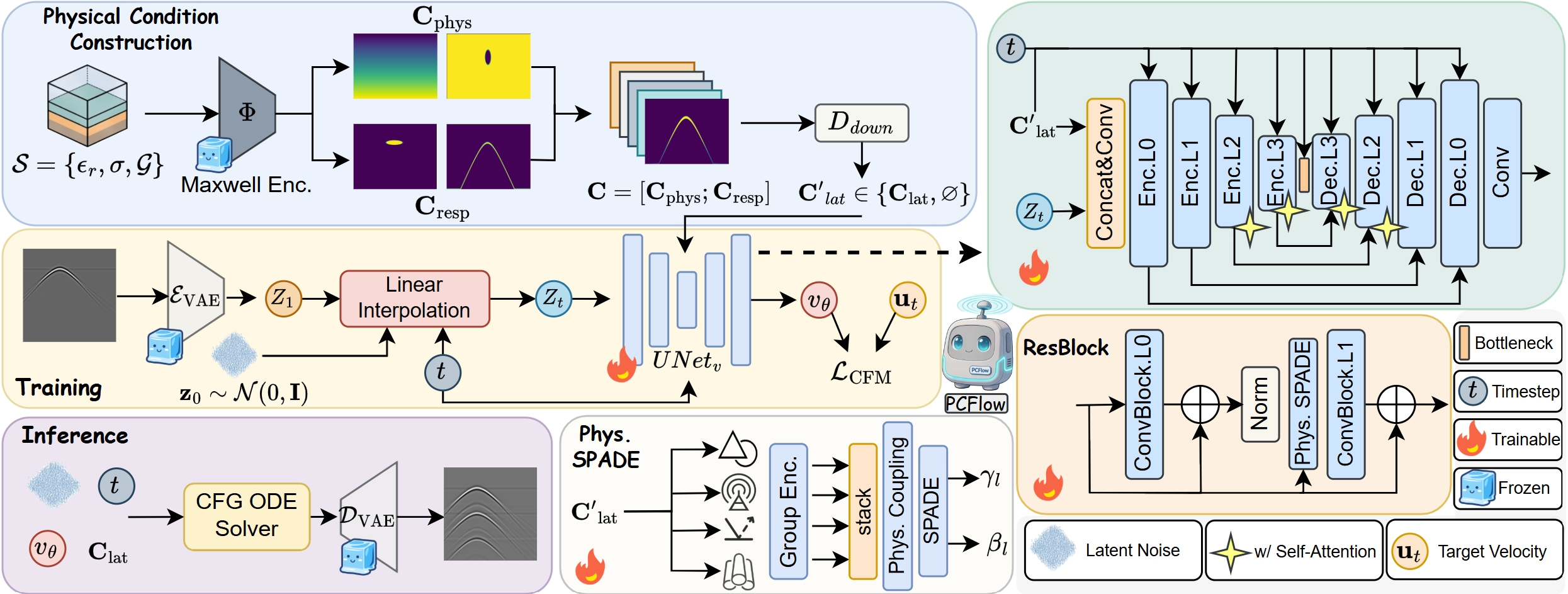}
    \caption{Overview of PCFlow. The Maxwell-informed encoder constructs physical and response-domain condition fields from the gprMax scene description, which are downsampled into latent conditions for conditional flow matching. During training, the velocity network predicts the latent transport direction from interpolated latents, while inference integrates the learned velocity field with CFG and decodes the final latent into a synthesized GPR B-scan. The right panels illustrate the conditional UNet backbone, Physics-SPADE modulation, and ResBlock design.}
    \label{fig:overview}
\end{figure*}

As shown in Fig.~\ref{fig:overview}, PCFlow separates GPR B-scan image synthesis into two coupled stages: physical condition construction and latent-space probability transport. The Maxwell-informed encoder provides interpretable physical and response-domain guidance, while the conditional velocity network transports Gaussian latent noise toward the VAE latent representation of the target B-scan~\cite{lipman2023flow,liu2023flow}. 

\subsection{Preliminary: Conditional Flow Matching}

Given $\mathbf z_0\sim\mathcal N(0,\mathbf I)$, $\mathbf z_1=\mathcal E_{\mathrm{VAE}}(I_{\mathrm B})$, and $t\sim\mathcal U(0,1)$, we use a linear probability path $\mathbf z_t=(1-t)\mathbf z_0+t\mathbf z_1$. The target velocity is $\mathbf u_t=\mathbf z_1-\mathbf z_0$. The conditional flow matching objective is
\begin{equation}
\label{eq:cfm_loss}
    \mathcal L_{\mathrm{CFM}}
    =
    \mathbb E
    \left[
    \left\|
    v_\theta(\mathbf z_t,t,\mathbf C_{\mathrm{lat}})
    -
    \mathbf u_t
    \right\|_2^2
    \right].
\end{equation}
Thus, training reduces to learning a conditional velocity field that transports Gaussian latent samples toward GPR image latents associated with the given physical condition.

The quality of the condition representation directly affects the ambiguity of this velocity regression problem. Under the squared loss, the Bayes optimal velocity is the conditional expectation
\begin{equation}
\label{eq:bayes_velocity_short}
    v^\star(\mathbf z_t,t,\mathbf C)
    =
    \mathbb E[\mathbf u_t\mid \mathbf z_t,t,\mathbf C].
\end{equation}
The remaining uncertainty of the transport direction can be measured by
\begin{equation}
    \mathcal R(\mathbf C)
    =
    \mathbb E
    \left[
    \operatorname{Tr}
    \operatorname{Var}(\mathbf u_t\mid\mathbf z_t,t,\mathbf C)
    \right].
\end{equation}
By the law of total variance, if $\mathbf{C}_b$ is an augmented representation such that $\sigma(\mathbf{C}_a) \subseteq \sigma(\mathbf{C}_b)$, the residual uncertainty is non-increasing, i.e., $\mathcal R(\mathbf{C}_b) \le \mathcal R(\mathbf{C}_a)$. A strict reduction in Bayes risk occurs whenever the additional Maxwell-informed priors provide non-redundant information that shifts the posterior mean velocity. This motivates our Maxwell-informed condition field, which exposes physical quantities directly coupled to the GPR response and guides the latent flow toward a more constrained and physically consistent transport path.

\subsection{Maxwell-Informed Condition Encoder}
\label{sec:maxwell_encoder}
The Maxwell-informed condition encoder $\Phi$ specifies how the physical simulation condition $\mathcal S$ is converted into the dense condition field used by conditional flow matching. Unlike a learnable black-box feature extractor, $\Phi$ is deterministic and is constructed from electromagnetic quantities that are known to influence GPR propagation and response geometry. Its role is to expose non-redundant physical information to the velocity network, thereby reducing the ambiguity of the conditional transport direction discussed in Sec.~\ref{sec:problem}.

Given the simulation condition $\mathcal S$, the encoder constructs
\begin{equation}
\label{eq:condition_decomposition}
    \mathbf C
    =
    \Phi(\mathcal S)
    =
    \operatorname{Concat}
    \left(
    \mathbf C_{\mathrm{phys}},
    \mathbf C_{\mathrm{resp}}
    \right),
\end{equation}
where $\mathbf C_{\mathrm{phys}}$ contains scene-domain physical fields and $\mathbf C_{\mathrm{resp}}$ contains radargram-domain response priors. These two groups are complementary. The scene-domain fields describe what is present in the subsurface model, including material properties, propagation cues, impedance or boundary discontinuities, and pipeline geometry. The response-domain priors describe where the dominant target-induced response is expected to appear in the B-scan, by projecting target geometry and background electromagnetic properties into the antenna-position--time coordinate system.

\subsubsection{From physical scene to response-domain geometry}

The need for $\mathbf C_{\mathrm{resp}}$ arises from the domain mismatch between the gprMax scene description and the B-scan image. A pipeline is represented as a localized geometric object in the physical scene, whereas its dominant radar response appears as a hyperbolic pattern in the radargram. Therefore, providing only scene-domain masks or material maps still requires the generator to infer the nonlinear mapping from target geometry and medium velocity to response travel time. We instead make this mapping explicit by constructing approximate response-domain priors from a simplified electromagnetic response model.

A compact way to describe the target-induced B-scan response is a delayed-wavelet integral~\cite{ORISTAGLIO2002448}:
\begin{equation}
\label{eq:bscan_master}
    I_s(y,x)
    =
    \int_{\Omega}
    W_s(x,\mathbf q)
    \chi_s(\mathbf q)
    \psi\left(y-T_s(x,\mathbf q)\right)
    d\mathbf q
    +
    r_s(y,x),
\end{equation}
where $s$ denotes the physical scene, $x$ is the antenna position, $y$ is the B-scan time or row coordinate, $\mathbf q$ denotes a scattering point in the subsurface, $\chi_s$ is the effective electromagnetic contrast, $T_s(x,\mathbf q)$ is the two-way travel time, $W_s$ summarizes geometric spreading, attenuation, antenna response, and reflection strength, and $r_s$ collects residual full-wave effects. This time-domain Born-type form shows that the response location in a radargram is governed by both material contrast and propagation time.

For the buried-pipeline setting, the dominant visible response is mainly associated with the earliest strong scattering component around the pipe. Eq.~\eqref{eq:bscan_master} can therefore be reduced to a shifted-wavelet response along a dominant two-way travel-time curve:
\begin{equation}
\label{eq:gpr_response_model}
    I_s(y,x)
    =
    A_s(x)\psi(y-\tau_s(x))
    +
    r_s(y,x),
\end{equation}
where $\tau_s(x)$ is the dominant travel-time curve and $A_s(x)$ represents the corresponding attenuation and reflection strength. This reduced model provides the first-order response geometry used by PCFlow. The conditional flow model then learns waveform details, interference, later propagation effects, and higher-order scattering patterns that are not specified by the analytical prior.

For a Ricker-like GPR pulse, the main lobe is locally sensitive to temporal shifts. In the relevant shift range, there exists a constant $c_\psi>0$ such that
\begin{equation}
\label{eq:shift_identifiable}
    \|\psi(\cdot-a)-\psi(\cdot-b)\|_2^2
    \ge
    c_\psi |a-b|^2 .
\end{equation}
Combining Eq.~\eqref{eq:gpr_response_model} with Eq.~\eqref{eq:shift_identifiable} gives a lower-bound-style relation between image-domain error and travel-time error:
\begin{equation}
\label{eq:curve_lower_bound}
    \|\hat I_s-I_s\|_2^2
    \gtrsim
    A_{\min}^2 c_\psi
    \|\hat\tau_s-\tau_s\|_2^2
    -
    O(\|r_s\|_2^2).
\end{equation}
This relation highlights the role of response-domain conditioning. A misplaced dominant travel-time curve induces a non-negligible image-domain discrepancy around the main reflection. By explicitly encoding an approximate $\tau_s(x)$ in $\mathbf C_{\mathrm{resp}}$, PCFlow reduces the burden of learning the nonlinear physical-parameter-to-response-geometry mapping from data alone. The velocity network can then focus on modeling the residual radargram appearance conditioned on this physically grounded geometry.

For a pipeline target with lateral center $x_c$ and burial depth $d$, we approximate the two-way travel time using a homogeneous-background propagation velocity $v_{\mathrm{bg}}=c_0/\sqrt{\epsilon_{r,\mathrm{bg}}}$:
\begin{equation}
\label{eq:twtt}
    \tau(x)
    \approx
    \frac{
    \sqrt{(x_{\mathrm{tx}}-x_c)^2+d^2}
    +
    \sqrt{(x_{\mathrm{rx}}-x_c)^2+d^2}
    }
    {c_0/\sqrt{\epsilon_{r,\mathrm{bg}}}},
\end{equation}
where $x_{\mathrm{tx}}$ and $x_{\mathrm{rx}}$ are the transmitter and receiver positions, $c_0$ is the speed of light, and $\epsilon_{r,\mathrm{bg}}$ is the background relative permittivity. Mapping $\tau(x)$ to the B-scan row coordinate yields an approximate hyperbola prior. The apex, attenuation, and target-strength priors are further derived from the travel-time minimum, propagation loss, and material or impedance contrast.

\subsubsection{Channel instantiation}

Based on the above design principle, the condition field used in this work contains 26 channels. The first two channels provide normalized radargram coordinates. The scene-domain physical field $\mathbf C_{\mathrm{phys}}$ contains material channels, propagation channels, boundary or impedance discontinuity channels, and explicit pipeline geometry channels. These channels describe wave velocity, attenuation, phase accumulation, reflection-related discontinuities, and target shape in the physical scene.

The response-domain field $\mathbf C_{\mathrm{resp}}$ is derived from the approximate two-way travel-time model. It contains six channels that encode the expected hyperbola heatmap, apex location, normalized travel-time curve, signed residual to the travel-time curve, attenuation along the response curve, and target-strength prior. These channels provide coarse but physically grounded anchors for the dominant B-scan response.

Although this work instantiates $\Phi$ for buried pipeline scenarios, the encoder should be understood as a general condition construction principle rather than a task-agnostic fixed channel design. For other GPR tasks, such as cavity detection, layered-medium identification, or multi-target recognition, the same principle can be applied by selecting task-specific physical priors and response mechanisms.

Table~\ref{tab:condition_encoder} summarizes the 26-channel condition field. When downsampling $\mathbf C$ to the VAE latent resolution, we use channel-aware pooling. Smooth dense fields are average-pooled, while sparse or localized fields are max-pooled. This average--max strategy preserves both continuous physical trends and sharp response-domain structures on the latent grid.

\begin{table*}[!t]
\centering
\caption{Composition of the Maxwell-informed condition field.}
\label{tab:condition_encoder}
\scriptsize
\setlength{\tabcolsep}{4pt}
\renewcommand{\arraystretch}{1.15}
\resizebox{\textwidth}{!}{
\begin{tabular}{
>{\centering\arraybackslash}p{0.09\textwidth}
p{0.15\textwidth}
>{\centering\arraybackslash}p{0.09\textwidth}
>{\centering\arraybackslash}p{0.09\textwidth}
p{0.55\textwidth}
}
\toprule
Group & Component & Channels & Pooling & Role \\
\midrule
Shared
& Coordinates
& 0--1
& Avg.
& Provide a radargram-domain spatial reference. \\
\midrule
\multirow{3}{*}{$\mathbf C_{\mathrm{phys}}$}
& Material
& 2--6
& Avg.
& Encode electromagnetic properties related to wave velocity, attenuation, and phase. \\

& Propagation
& 7--11
& Avg. / Max.
& Encode depth-wise propagation cues and reflection-related discontinuities. \\

& Geometry
& 12--19
& Avg. / Max.
& Encode pipe support, boundary, orientation, size, and location. \\
\midrule
$\mathbf C_{\mathrm{resp}}$
& Response priors
& 20--25
& Avg. / Max.
& Anchor the expected hyperbola, apex, travel-time trend, attenuation, and response strength. \\

\bottomrule
\end{tabular}
}
\end{table*}

\subsection{Conditional Velocity Network and Sampling}

The conditional velocity field $v_\theta(\mathbf z_t,t,\mathbf C_{\mathrm{lat}})$ is parameterized by a UNet-style network~\cite{ronneberger2015u,rombach2022high}. The input latent state and the downsampled condition field are first concatenated and projected by a convolutional stem:
\begin{equation}
    \mathbf h_0 =
    \operatorname{Conv}_{3\times3}
    \left(\operatorname{Concat}[\mathbf z_t,\mathbf C_{\mathrm{lat}}]\right).
\end{equation}
The timestep $t$ is encoded by a sinusoidal embedding followed by a multilayer perceptron, and is injected into each residual block through an additive time projection~\cite{ho2020denoising,he2016deep}. The network follows an encoder--bottleneck--decoder architecture with skip connections. At each resolution level, two residual blocks are used before downsampling or upsampling, and self-attention is applied at selected low-resolution feature maps to capture long-range response dependencies~\cite{vaswani2017attention}.

\subsubsection{SPADE-based physical modulation}

To provide multi-scale physical conditioning, we build a condition pyramid from $\mathbf C_{\mathrm{lat}}$ and feed the resized condition field to residual blocks at the corresponding resolution. Beyond input concatenation, we use SPADE-based modulation~\cite{park2019semantic} to inject dense physical conditions into intermediate velocity features:
\begin{equation}
    \gamma_l,\beta_l=\Psi_l(\mathbf C_l), \qquad
    \operatorname{Mod}(\mathbf h_l,\mathbf C_l)
    =
    (1+\gamma_l)\odot \operatorname{Norm}(\mathbf h_l)+\beta_l ,
\end{equation}
where $\mathbf C_l$ is the condition field resized to the feature resolution of layer $l$, and $\Psi_l$ predicts spatially varying affine parameters.

This design is motivated by the structure of GPR B-scans. A radargram is a single-channel grayscale response image whose physical semantics are mainly carried by local amplitude variations, polarity changes, and hyperbolic reflection patterns. Since these structures are tied to antenna position and travel time, conditioning only at the network input may be insufficient. SPADE-based modulation repeatedly injects location-dependent physical guidance at multiple resolutions, helping the velocity network preserve response geometry while learning residual scattering and appearance details.

We consider two variants. PCFlow (PL) uses a plain all-channel modulation network, where $\Psi_l$ predicts $\gamma_l$ and $\beta_l$ directly from all condition channels. This variant allows flexible cross-channel interaction and often provides strong image fidelity when the test distribution is close to the training data. PCFlow (GR), or Physics-SPADE, partitions the condition channels into physics-related groups, such as material, propagation, geometry, reflection-related, and response-prior groups. These groups are encoded separately and then fused to predict the modulation parameters. This grouped design is introduced to better exploit the dense and heterogeneous physical condition field under distribution shifts, where material attenuation, response strength, and visible geometry may change jointly. By reducing early mixing between heterogeneous cues, PCFlow (GR) is expected to preserve geometry-relevant and response-prior information more robustly in OOD settings.

\subsubsection{Classifier-free guided sampling}

To enable classifier-free guidance~\cite{ho2021classifierfree}, the condition field is randomly dropped during training with probability $p_{\mathrm{drop}}$, so that the same network learns both conditional and unconditional velocity estimates. At sampling time, we use the guided velocity
\begin{equation}
\label{eq:cfg_velocity}
    v_{\mathrm{cfg}}
    =
    v_\theta(\mathbf z_t,t,\varnothing)
    +
    s\left[
    v_\theta(\mathbf z_t,t,\mathbf C_{\mathrm{lat}})
    -
    v_\theta(\mathbf z_t,t,\varnothing)
    \right],
\end{equation}
where $s\ge1$ is the guidance scale. We employ CFG to amplify condition-specific transport, yielding deterministic physical outputs aligned with the encoded simulation condition. Starting from $\mathbf z_0\sim\mathcal N(0,\mathbf I)$, we integrate $d\mathbf z_t/dt=v_{\mathrm{cfg}}(\mathbf z_t,t,\mathbf C_{\mathrm{lat}})$ from $t=0$ to $t=1$, and decode the final latent with $\mathcal D_{\mathrm{VAE}}$.

\section{Experiments}

\subsection{Dataset and Splits}

We construct a synthetic buried-pipeline GPR B-scan dataset using gprMax~\cite{warren2016gprmax} full-wave electromagnetic simulation. The dataset contains approximately 800 paired samples, each consisting of a gprMax configuration file and the corresponding direct-wave-removed B-scan. The data cover diverse soil backgrounds and pipeline configurations, including dry sand, wet sand, wet soil, and high-conductivity dry clay, as well as steel pipes and PVC pipes filled with air or water. Pipe radii, lateral positions, and burial depths are varied to produce different response geometries and reflection strengths.

Samples from dry sand, wet sand, and wet soil are used as the in-distribution (ID) subset, while the high-conductivity dry-clay cases are held out as the OOD stress test. To avoid leakage from horizontally shifted variants of the same physical scene, we perform group-aware splitting based on pipe material, filling, radius, soil properties, and vertical position. The final split contains 443 training samples, 88 validation samples, 70 ID test samples, and 199 OOD test samples, as summarized in Table~\ref{tab:dataset_split_summary}.
\begin{table*}[!t]
\centering
\caption{Dataset split summary. Each soil pair denotes $(\epsilon_r,\sigma)$, where $\sigma$ is in S/m. Here $d=1.0-y_c$ denotes the metadata burial depth, assuming the soil-surface coordinate is $y=1.0~\mathrm{m}$.}
\label{tab:dataset_split_summary}
\scriptsize
\setlength{\tabcolsep}{4pt}
\renewcommand{\arraystretch}{1.12}
\resizebox{\textwidth}{!}{
\begin{tabular}{lcllll}
\toprule
Split & \#Samples & Soil $(\epsilon_r,\sigma)$ & Target & Radius $r$ (m) & Position (m) \\
\midrule
Train
& 443
& $(4,0.001)$, $(10,0.005)$, $(20,0.02)$
& Steel, PVC (air), PVC (water)
& $\{0.03,0.05,0.08,0.10\}$
& $x_c\in[1.00,2.99]$, $y_c\in[0.25,0.80]$, $d\in[0.20,0.75]$ \\

Val
& 88
& $(4,0.001)$, $(10,0.005)$, $(20,0.02)$
& Steel, PVC (air), PVC (water)
& $\{0.03,0.05,0.08,0.10\}$
& $x_c\in[1.00,2.99]$, $y_c\in[0.25,0.80]$, $d\in[0.20,0.75]$ \\

Test-ID
& 70
& $(4,0.001)$, $(10,0.005)$, $(20,0.02)$
& Steel, PVC (air), PVC (water)
& $\{0.03,0.05,0.08,0.10\}$
& $x_c\in[1.04,2.99]$, $y_c\in[0.25,0.80]$, $d\in[0.20,0.75]$ \\

OOD
& 199
& $(5,0.05)$
& Steel, PVC (air), PVC (water)
& $\{0.03,0.05,0.08,0.10\}$
& $x_c\in[1.01,2.99]$, $y_c\in[0.26,0.79]$, $d\in[0.21,0.74]$ \\
\bottomrule
\end{tabular}
}
\end{table*}

\subsection{Baselines}

We compare PCFlow with representative baselines from three categories. All baselines are trained on the same split, with hyperparameters selected according to validation performance to ensure a fair comparison.

\textbf{GAN-based image-to-image translation:} Pix2Pix~\cite{isola2017image} is used as the GAN-based baseline and takes the same dense physical condition field as input.

\textbf{Conditional and controllable diffusion:} For conditional DDPM~\cite{ho2020denoising}, discrete physical attributes are encoded by one-hot lookup tables, while continuous parameters are normalized and concatenated into a global condition vector injected into the UNet. The physics-conditioned DDPM adds a hyperbola-prior-weighted reconstruction constraint to the standard denoising loss~\cite{bastek2025physics}. DDBM~\cite{zhou2024denoising} renders physical parameters into a three-channel spatial prior image, including approximate hyperbola shape, apex location, and material-soil attributes. The ControlNet-style~\cite{zhang2023adding} baseline maps the dense condition field to a single-channel control image through a learnable $1\times1$ convolution.

\textbf{Text-guided latent diffusion:} Stable Diffusion 2.1 baselines use fixed prompt templates converted from physical parameters~\cite{rombach2022high}. We evaluate zero-shot generation and fine-tuning variants, including LoRA tuning~\cite{hu2022lora} of the UNet, DataDream-style joint LoRA of UNet and text encoder~\cite{kim2024datadream}, and full UNet fine-tuning.

\begin{table*}[!t]
\centering
\caption{Geometry-aware evaluation on ID and OOD test sets. Opening error is scaled by $10^{-3}$.}
\label{tab:geometry_main}
\scriptsize
\resizebox{\textwidth}{!}{
\begin{tabular}{llccccc|ccccc}
\toprule
\multirow{2}{*}{Category} & \multirow{2}{*}{Method}
& \multicolumn{5}{c|}{ID test}
& \multicolumn{5}{c}{OOD test} \\
\cmidrule(lr){3-7}\cmidrule(lr){8-12}
& & Apex-x Err. $\downarrow$ & Apex-y Err. $\downarrow$ & Curve Err. $\downarrow$
& Opening Err. $\downarrow$ & IoU $\uparrow$
& Apex-x Err. $\downarrow$ & Apex-y Err. $\downarrow$ & Curve Err. $\downarrow$
& Opening Err. $\downarrow$ & IoU $\uparrow$ \\
\midrule
GAN / I2I
& Pix2Pix~\cite{isola2017image}
& 12.86 & 11.39 & 16.58 & 1.29 & 0.27
& 13.98 & 50.06 & 52.94 & 3.49 & 0.00 \\
\midrule
\multirow{4}{*}{Conditional / controllable diffusion}
& DDBM~\cite{zhou2024denoising}
& 27.58 & 54.42 & 53.08 & 3.71 & 0.00
& 37.21 & 57.52 & 58.03 & 4.11 & 0.00 \\
& DDPM~\cite{ho2020denoising}
& 43.94 & 67.44 & 103.65 & 4.13 & 0.00
& 37.33 & 70.87 & 111.87 & 4.05 & 0.00 \\
& Phy-DDPM~\cite{bastek2025physics}
& 59.38 & 55.87 & 92.24 & 2.29 & 0.00
& 49.57 & 70.72 & 148.09 & 3.56 & 0.00 \\
& ControlNet~\cite{zhang2023adding}
& 58.80 & 80.52 & 96.10 & 5.24 & 0.00
& 57.82 & 72.55 & 111.75 & 7.16 & 0.00 \\
\midrule
\multirow{4}{*}{Text-guided diffusion}
& SD Zero-shot~\cite{rombach2022high}
& 29.54 & 90.67 & 138.81 & 11.35 & 0.00
& 33.36 & 62.94 & 127.74 & 8.73 & 0.00 \\
& SD-LoRA~\cite{hu2022lora}
& 65.50 & 69.20 & 70.95 & 4.67 & 0.01
& 68.00 & 79.64 & 107.34 & 3.34 & 0.00 \\
& DataDream~\cite{kim2024datadream}
& 66.57 & 67.58 & 83.83 & 3.70 & 0.02
& 65.97 & 86.38 & 117.64 & 2.81 & 0.00 \\
& SD-FFT~\cite{rombach2022high}
& 60.51 & 66.08 & 67.84 & 4.77 & 0.00
& 65.03 & 61.04 & 87.36 & 3.05 & 0.00 \\
\midrule
\multirow{2}{*}{Ours}
& PCFlow (PL)
& \textbf{0.55} & \textbf{0.68} & \textbf{1.01} & 0.16 & \textbf{0.91}
& 1.53 & 7.12 & 6.80 & \textbf{0.41} & 0.69 \\
& PCFlow (GR)
& 0.62 & 0.82 & \textbf{1.01} & \textbf{0.14} & \textbf{0.91}
& \textbf{1.10} & \textbf{2.94} & \textbf{3.57} & 0.60 & \textbf{0.74} \\
\bottomrule
\end{tabular}
}
\end{table*}

\begin{table*}[!t]
\centering
\caption{Image fidelity evaluation on ID and OOD test sets.}
\label{tab:image_main}
\scriptsize
\resizebox{\textwidth}{!}{
\begin{tabular}{llccc|ccc}
\toprule
\multirow{2}{*}{Category} & \multirow{2}{*}{Method}
& \multicolumn{3}{c|}{ID test}
& \multicolumn{3}{c}{OOD test} \\
\cmidrule(lr){3-5}\cmidrule(lr){6-8}
& & PSNR $\uparrow$ & SSIM $\uparrow$ & LPIPS $\downarrow$
& PSNR $\uparrow$ & SSIM $\uparrow$ & LPIPS $\downarrow$ \\
\midrule
GAN / I2I
& Pix2Pix~\cite{isola2017image}
& $24.37{\pm}2.06$ & $0.8636{\pm}0.0849$ & $0.2259{\pm}0.0588$
& $19.81{\pm}1.93$ & $0.8066{\pm}0.0634$ & $0.4560{\pm}0.1122$ \\
\midrule
\multirow{4}{*}{Conditional / controllable diffusion}
& DDBM~\cite{zhou2024denoising}
& $23.56{\pm}1.61$ & $0.7535{\pm}0.0754$ & $0.5514{\pm}0.0721$
& $18.94{\pm}1.62$ & $0.7237{\pm}0.0489$ & $0.6138{\pm}0.0481$ \\
& DDPM~\cite{ho2020denoising}
& $17.00{\pm}4.78$ & $0.5054{\pm}0.3198$ & $0.6299{\pm}0.2265$
& $16.83{\pm}5.40$ & $0.5228{\pm}0.3121$ & $0.6395{\pm}0.2232$ \\
& Phy-DDPM~\cite{bastek2025physics}
& $23.51{\pm}2.01$ & $0.7308{\pm}0.1920$ & $0.5089{\pm}0.1734$
& $19.80{\pm}2.78$ & $0.7588{\pm}0.1689$ & $0.5075{\pm}0.1708$ \\
& ControlNet~\cite{zhang2023adding}
& $21.90{\pm}1.87$ & $0.8219{\pm}0.0638$ & $0.4525{\pm}0.0518$
& $18.28{\pm}3.11$ & $0.8215{\pm}0.0492$ & $0.4800{\pm}0.0825$ \\
\midrule
\multirow{4}{*}{Text-guided diffusion}
& SD Zero-shot~\cite{rombach2022high}
& $10.01{\pm}2.31$ & $0.1539{\pm}0.1267$ & $0.9543{\pm}0.0687$
& $9.63{\pm}2.31$ & $0.1639{\pm}0.1346$ & $0.9557{\pm}0.0760$ \\
& SD-LoRA~\cite{hu2022lora}
& $22.65{\pm}1.64$ & $0.8474{\pm}0.0702$ & $0.3363{\pm}0.0762$
& $18.87{\pm}1.86$ & $0.8485{\pm}0.0534$ & $0.3547{\pm}0.0780$ \\
& DataDream~\cite{kim2024datadream}
& $22.86{\pm}1.79$ & $0.8621{\pm}0.0771$ & $0.3108{\pm}0.0755$
& $18.88{\pm}2.20$ & $0.8691{\pm}0.0510$ & $0.3129{\pm}0.0744$ \\
& SD-FFT~\cite{rombach2022high}
& $21.65{\pm}2.94$ & $0.7848{\pm}0.1841$ & $0.3334{\pm}0.1186$
& $18.34{\pm}2.71$ & $0.8075{\pm}0.1469$ & $0.3245{\pm}0.1187$ \\
\midrule
\multirow{2}{*}{Ours}
& PCFlow (PL)
& $\textbf{33.31}{\pm}5.65$ & $\textbf{0.9611}{\pm}0.0538$ & $\textbf{0.0398}{\pm}0.0381$
& $19.82{\pm}2.06$ & $\textbf{0.9028}{\pm}0.0342$ & $\textbf{0.1875}{\pm}0.0798$ \\
& PCFlow (GR)
& $32.42{\pm}5.17$ & $0.9571{\pm}0.0616$ & $0.0842{\pm}0.0509$
& $\textbf{19.89}{\pm}2.57$ & $0.8850{\pm}0.0424$ & $0.1939{\pm}0.0534$ \\
\bottomrule
\end{tabular}
}
\end{table*}

\subsection{Implementation Details}

PCFlow uses a pretrained SDXL VAE~\cite{podell2024sdxl} to encode and decode B-scans in latent space, with the posterior mean used as the target latent. The conditional velocity network is a UNet-style model with 4 latent channels, a time embedding dimension of 128, a base width of 128, and a dropout rate of 0.1. The Maxwell-informed condition encoder uses the simulation grid spacing $(0.0039, 0.0012)$ and a center frequency of $400~\mathrm{MHz}$.

All models are trained for 10k iterations with batch size 8 and learning rate $1\times10^{-4}$. The condition dropout probability is set to 0.1 for classifier-free guidance. At inference, we use the Heun solver with 50 sampling steps and guidance scale 2.5. The EMA checkpoint is used for evaluation.

\begin{figure*}[!t]
    \centering
    \includegraphics[width=0.96\textwidth]{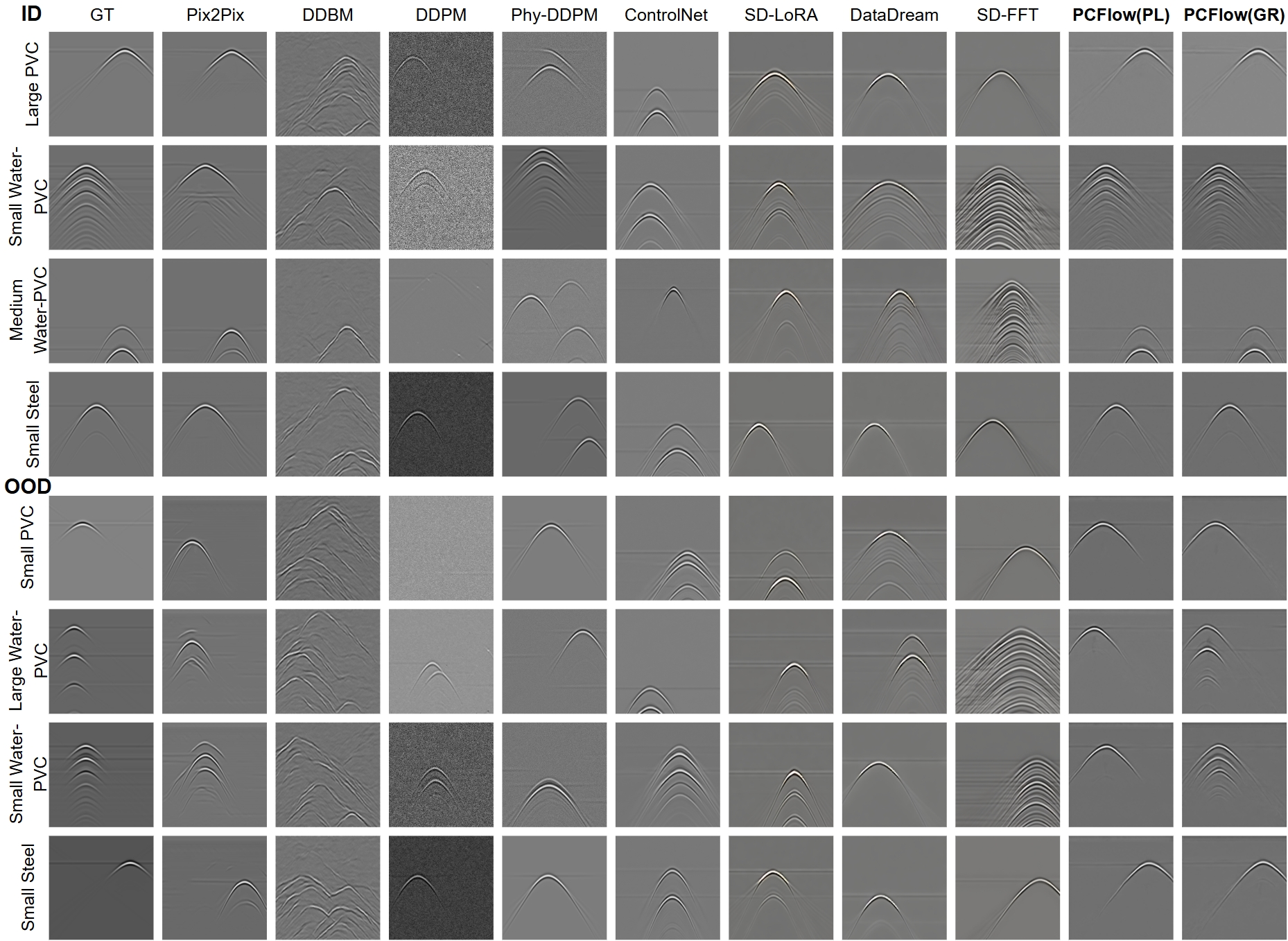}
    \caption{Qualitative comparison on ID and OOD test samples. PCFlow generates more physically consistent hyperbolic responses than general generative baselines. While PCFlow (PL) achieves strong visual fidelity, PCFlow (GR) shows better OOD response geometry, especially under high-conductivity distribution shifts.}
    \label{fig:qualitative_comparison}
\end{figure*}

\subsection{Evaluation Metrics and Results}

\subsubsection{Metrics}
We evaluate both image fidelity and physical response consistency. Image fidelity is measured by PSNR, SSIM~\cite{wang2004image}, and LPIPS~\cite{zhang2018unreasonable}. Since these image-level metrics cannot fully reflect whether the generated radar response is physically aligned with the target, we further introduce training-free geometry-aware metrics based on the dominant early hyperbolic response.

For each ground-truth B-scan $I$ and generated B-scan $\hat I$, we extract the dominant response masks $M$ and $\hat M$, as well as their ridge curves $r(x)$ and $\hat r(x)$, using robust intensity normalization, background suppression, smoothing, and connected-component filtering. The apexes of the ground-truth and generated responses are denoted by $(x_a,y_a)$ and $(\hat x_a,\hat y_a)$, respectively. The horizontal and vertical \textbf{apex errors} are defined as
\begin{equation}
\label{eq:apex_error}
    E_{\mathrm{apex}\text{-}x}=|\hat x_a-x_a|,
    \qquad
    E_{\mathrm{apex}\text{-}y}=|\hat y_a-y_a|.
\end{equation}

The \textbf{curve error} measures the mean absolute ridge displacement over the shared valid lateral support $\Omega$:
\begin{equation}
\label{eq:curve_error}
    E_{\mathrm{curve}}
    =
    \frac{1}{|\Omega|}
    \sum_{x\in\Omega}
    |\hat r(x)-r(x)|.
\end{equation}

To evaluate the aperture shape of the hyperbolic response, each ridge is locally fitted around the apex by a quadratic curve $y=a(x-x_a)^2+b(x-x_a)+c$. The \textbf{opening error} is defined as
\begin{equation}
\label{eq:opening_error}
    E_{\mathrm{open}}=|\hat a-a|.
\end{equation}

We also report the response-mask IoU~\cite{everingham2010pascal} to measure the overlap between $M$ and $\hat M$. Apex and curve errors are reported in pixels, opening errors are reported in $10^{-3}$ units, and higher IoU indicates better response-region alignment. These geometry-aware metrics jointly evaluate the localization, depth, curve consistency, and aperture shape of the dominant early hyperbola.

\begin{figure*}[!t]
    \centering
    \includegraphics[width=0.95\textwidth]{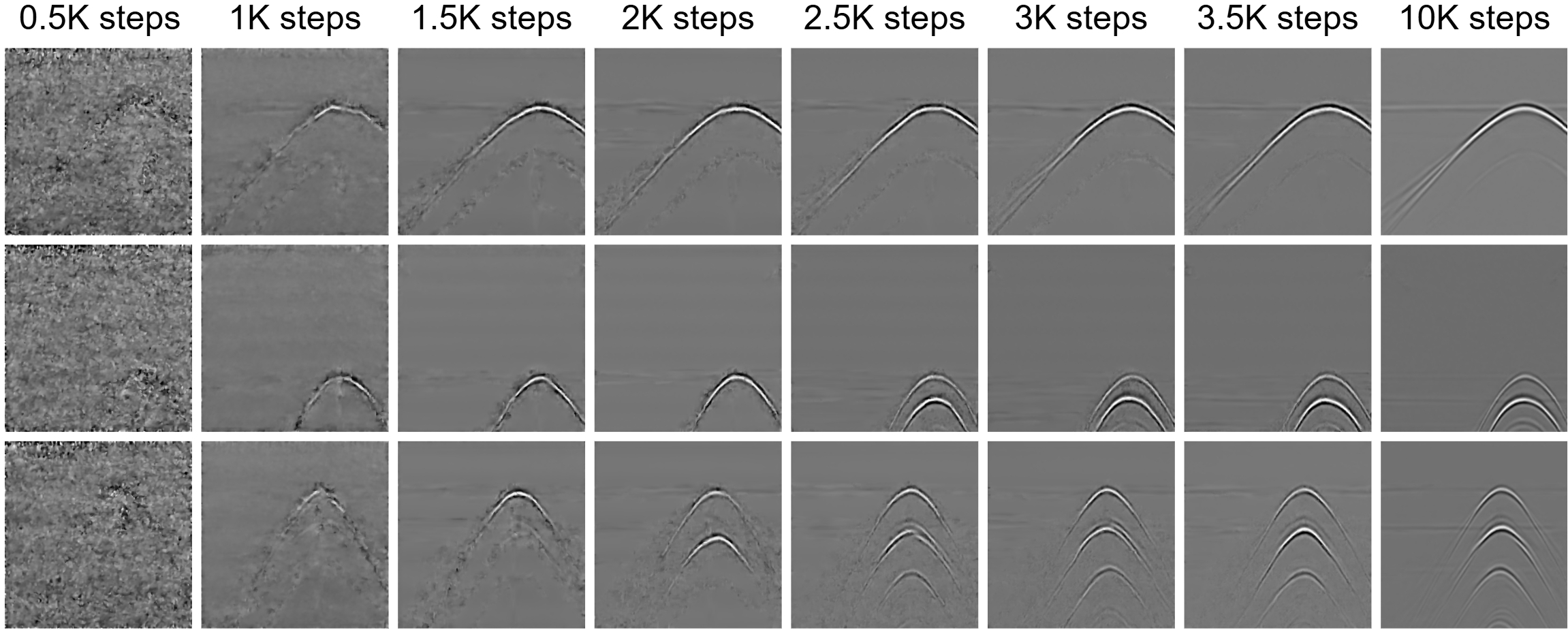}
    \caption{Training evolution of PCFlow on fixed validation samples. Weak target-response structures appear at 0.5K iterations, and the dominant hyperbolic response becomes clear by 1K iterations. Around 2K--2.5K iterations, secondary reflection branches and multi-return structures are gradually formed. Later iterations mainly improve waveform sharpness, background smoothness, and overall radargram fidelity.}
    \label{fig:training_evolution}
\end{figure*}

\subsubsection{Main results}
Table~\ref{tab:geometry_main} reports geometry-aware evaluation results on both ID and OOD test sets. General-purpose generative baselines, including GAN-based image translation, conditional diffusion, controllable diffusion, and text-guided diffusion, struggle to preserve the target-induced response geometry. Many baselines produce near-zero mask IoU, indicating that the generated hyperbolic response is misplaced or physically invalid. In contrast, PCFlow achieves substantially lower apex and curve errors and much higher mask IoU on both ID and OOD test sets, as also shown qualitatively in Fig.~\ref{fig:qualitative_comparison}.

Table~\ref{tab:image_main} reports image-level quality metrics. PCFlow also achieves strong image fidelity, especially on the ID test set. More importantly, the comparison between Tables~\ref{tab:geometry_main} and~\ref{tab:image_main} shows that conventional image metrics are not sufficient for evaluating physics-conditioned GPR B-scan image synthesis. Some baselines obtain moderate PSNR or SSIM but fail to produce physically meaningful response regions. Fig.~\ref{fig:training_evolution} further visualizes the training evolution of PCFlow on fixed validation samples. The dominant response geometry is established in the early training stage, while later iterations mainly refine secondary returns and radargram appearance.

\begin{table*}[!t]
\centering
\caption{Ablation study on geometry-aware metrics. Opening error is scaled by $10^{-3}$. Lower is better for error metrics, and higher is better for Mask IoU.}
\label{tab:ablation_geometry}
\scriptsize
\resizebox{\textwidth}{!}{
\begin{tabular}{lccccc|ccccc}
\toprule
\multirow{2}{*}{Variant}
& \multicolumn{5}{c|}{ID test}
& \multicolumn{5}{c}{OOD test} \\
\cmidrule(lr){2-6}\cmidrule(lr){7-11}
& Apex-x Err. $\downarrow$ & Apex-y Err. $\downarrow$ & Curve Err. $\downarrow$
& Opening Err. $\downarrow$ & IoU $\uparrow$
& Apex-x Err. $\downarrow$ & Apex-y Err. $\downarrow$ & Curve Err.$\downarrow$
& Opening Err. $\downarrow$ & IoU $\uparrow$ \\
\midrule
PCFlow (GR)
& 0.62 & 0.82 & \textbf{1.01} & \textbf{0.14} & \textbf{0.91}
& 1.10 & \textbf{2.94} & \textbf{3.57} & 0.60 & \textbf{0.74} \\
PCFlow (PL)
& \textbf{0.55} & \textbf{0.68} & \textbf{1.01} & 0.16 & \textbf{0.91}
& 1.53 & 7.12 & 6.80 & \textbf{0.41} & 0.69 \\
PCFlow w/o SPADE
& 1.13 & 0.91 & 1.32 & 0.23 & 0.89
& 1.95 & 7.68 & 6.18 & 0.74 & 0.60 \\
PCFlow w/o response priors
& 1.62 & 1.11 & 2.02 & 0.27 & 0.85
& 1.67 & 96.83 & 106.02 & 4.73 & 0.00 \\
PCFlow w/ material only
& 1.78 & 1.52 & 2.17 & 0.25 & 0.86
& 2.12 & 102.87 & 115.31 & 5.43 & 0.00 \\
PCFlow w/ response priors only
& 1.29 & 1.16 & 1.60 & 0.17 & 0.89
& \textbf{1.03} & 4.94 & 4.64 & 0.48 & \textbf{0.74} \\
PCFlow w/ geometry only
& 1.93 & 52.99 & 62.98 & 3.80 & 0.00
& 2.36 & 45.56 & 51.68 & 2.47 & 0.00 \\
\bottomrule
\end{tabular}
}
\end{table*}

\begin{table*}[!t]
\centering
\caption{Ablation study on image fidelity metrics. Higher is better for PSNR and SSIM, and lower is better for LPIPS.}
\label{tab:ablation_image}
\scriptsize
\resizebox{\textwidth}{!}{
\begin{tabular}{lccc|ccc}
\toprule
\multirow{2}{*}{Variant}
& \multicolumn{3}{c|}{ID test}
& \multicolumn{3}{c}{OOD test} \\
\cmidrule(lr){2-4}\cmidrule(lr){5-7}
& PSNR $\uparrow$ & SSIM $\uparrow$ & LPIPS $\downarrow$
& PSNR $\uparrow$ & SSIM $\uparrow$ & LPIPS $\downarrow$ \\
\midrule
PCFlow (GR)
& $32.42{\pm}5.17$ & $0.9571{\pm}0.0616$ & $0.0842{\pm}0.0509$
& $19.89{\pm}2.57$ & $0.8850{\pm}0.0424$ & $0.1939{\pm}0.0534$ \\
PCFlow (PL)
& $\textbf{33.31}{\pm}5.65$ & $\textbf{0.9611}{\pm}0.0538$ & $\textbf{0.0398}{\pm}0.0381$
& $19.82{\pm}2.06$ & $\textbf{0.9028}{\pm}0.0342$ & $\textbf{0.1875}{\pm}0.0798$ \\
PCFlow w/o SPADE
& $31.84{\pm}4.61$ & $0.9572{\pm}0.0501$ & $0.0453{\pm}0.0428$
& $19.69{\pm}2.33$ & $0.8796{\pm}0.0352$ & $0.2648{\pm}0.0704$ \\
PCFlow w/o response priors
& $29.98{\pm}4.98$ & $0.9391{\pm}0.0701$ & $0.0462{\pm}0.0410$
& $20.17{\pm}2.39$ & $0.8650{\pm}0.0381$ & $0.3674{\pm}0.0529$ \\
PCFlow w/ material only
& $29.14{\pm}4.37$ & $0.9367{\pm}0.0689$ & $0.0479{\pm}0.0464$
& $\textbf{20.49}{\pm}2.54$ & $0.8690{\pm}0.0306$ & $0.3573{\pm}0.0399$ \\
PCFlow w/ response priors only
& $30.39{\pm}6.93$ & $0.9307{\pm}0.0839$ & $0.0957{\pm}0.1160$
& $20.38{\pm}2.67$ & $0.9004{\pm}0.0499$ & $0.2029{\pm}0.0791$ \\
PCFlow w/ geometry only
& $24.14{\pm}1.84$ & $0.8850{\pm}0.0569$ & $0.2708{\pm}0.1346$
& $19.78{\pm}2.20$ & $0.8838{\pm}0.0381$ & $0.2723{\pm}0.0822$ \\
\bottomrule
\end{tabular}
}
\end{table*}

\begin{figure*}[!t]
    \centering
    \includegraphics[width=0.95\textwidth]{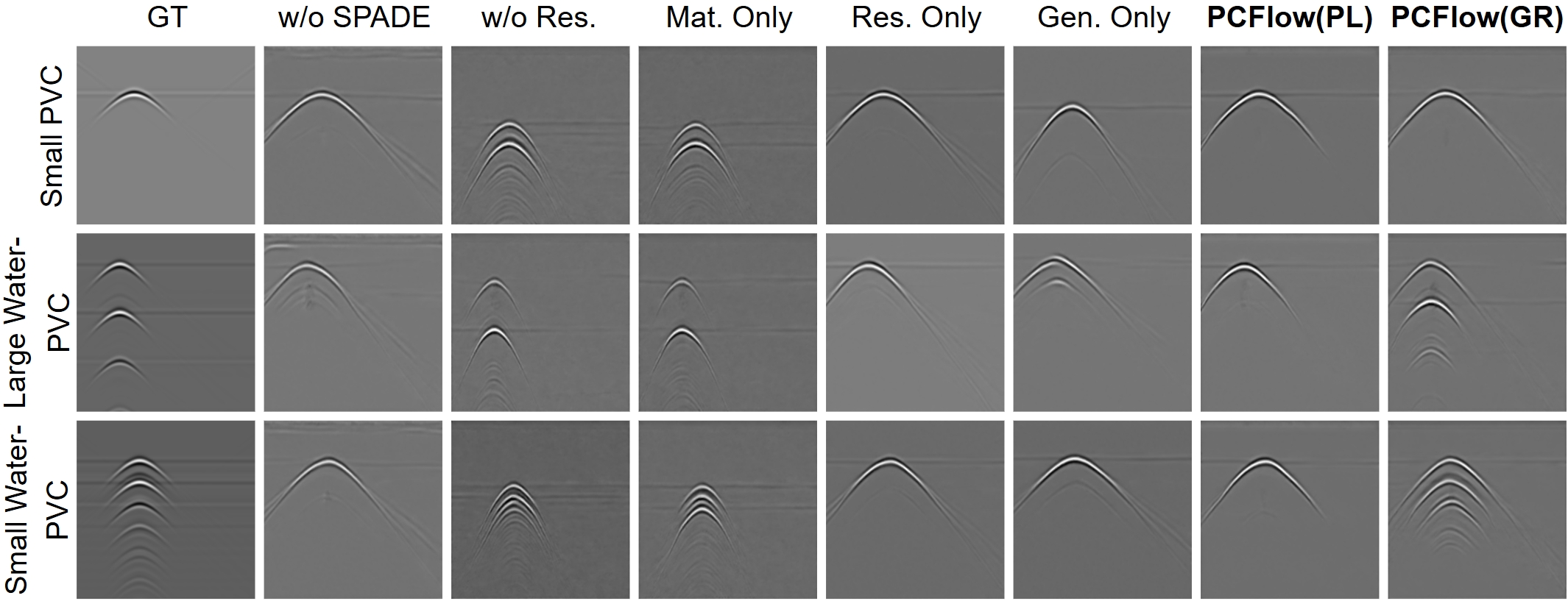}
    \caption{Qualitative ablation comparison on OOD test samples. Mat. Only, w/o Res., w/o SPADE, Res. Only, and Geo. Only denote PCFlow w/ material only, w/o response priors, w/o SPADE modulation, w/ response priors only, and w/ geometry only, respectively. The results show that response-domain priors are crucial for preserving OOD hyperbola geometry, while PCFlow (GR) yields more stable physically aligned responses.}
    \label{fig:ablation_ood}
\end{figure*}

\section{Ablation Analysis}

We further analyze the contributions of the physical condition fields and the condition modulation strategy. Besides the two complete PCFlow variants, PCFlow (GR) with grouped modulation and PCFlow (PL) with plain all-channel modulation, we consider five ablated settings. PCFlow w/o SPADE removes spatially adaptive modulation and only uses input concatenation. PCFlow w/o response priors removes $\mathbf C_{\mathrm{resp}}$ and keeps the first 20 scene-domain physical channels. PCFlow w/ material only keeps the coordinate, material, electromagnetic propagation, and edge-related channels, namely channels 0--11, while removing explicit pipe geometry and response-domain priors. PCFlow w/ response priors only keeps the coordinate channels and the six response-domain prior channels, namely channels 0--1 and 20--25. PCFlow w/ geometry only keeps the coordinate channels and explicit pipe geometry channels, namely channels 0--1 and 12--19, while removing material, propagation, boundary, and response-domain prior fields.
Tables~\ref{tab:ablation_geometry} and~\ref{tab:ablation_image} show that the Maxwell-informed condition field is the main source of PCFlow's physical consistency. Removing response-domain priors causes severe OOD geometry failure: the curve error increases from $3.57$ to $106.02$, and Mask IoU drops from $0.74$ to $0$. Similarly, the material-only variant maintains reasonable OOD PSNR but fails to localize the hyperbolic response, indicating that material and propagation fields alone are insufficient to determine response geometry under distribution shift.

The response-prior-only variant achieves competitive OOD geometry consistency, showing that response-domain priors provide strong geometric anchors, as also illustrated in Fig.~\ref{fig:ablation_ood}. However, its image fidelity is weaker than the full models, suggesting that material, propagation, and boundary cues are still needed to recover detailed radargram appearance. In contrast, the geometry-only variant performs poorly, confirming that explicit target shape and location maps in the scene domain are not sufficient by themselves. They need to be projected into the radargram domain through travel-time and response priors to provide effective geometric guidance for B-scan image synthesis.

The modulation ablation shows that PCFlow (PL) gives the strongest image fidelity, while PCFlow (GR) achieves the best OOD curve consistency. Removing SPADE weakens both geometry and image metrics. Overall, the results indicate that PCFlow's main advantage comes from the Maxwell-informed condition field, especially the response-domain priors, rather than from a single architectural choice.

\section{Discussion}
\paragraph{Physical Reliability vs. Visual Fidelity}
The experimental results suggest that visually plausible GPR B-scan image synthesis is not necessarily physically reliable. Several baseline models achieve moderate image-level scores, such as PSNR and SSIM, but their response-mask IoU remains close to zero. This indicates that the target-induced hyperbola is either severely misplaced or missing entirely. This stark contrast confirms the critical need for geometry-aware evaluation metrics when generative models are deployed for physics-conditioned radargram synthesis. Furthermore, the two PCFlow variants exhibit complementary behaviors under distribution shifts. PCFlow (PL) tends to provide stronger in-distribution image sharpness, whereas PCFlow (GR) yields more stable response geometry in high-conductivity OOD scenarios. This trade-off implies that grouped physical modulation is highly beneficial when heterogeneous condition channels, including attenuation, material contrast, and response priors, vary jointly outside the training distribution.

\paragraph{Nature of the Maxwell-Informed Guidance}
The proposed response-domain priors should be interpreted as a physically informed, soft guiding framework rather than an analytical substitute for full-wave electromagnetic simulation. Instead of hard-coding the final radargram, this paradigm projects scene-domain physical parameters into radargram-aligned response anchors, exposing the dominant travel-time geometry and coarse amplitude trends to the generative model. Meanwhile, highly complex physical mechanisms, such as residual wave scattering, multi-path interference, fine waveform phase changes, and antenna-related coupling effects, are adaptively learned by the conditional velocity network from paired gprMax simulation data. Essentially, the condition field acts as a task-specific instantiation of a general condition-construction principle, guiding the transport pathway in the VAE latent space toward physically consistent manifolds while preserving the generative capacity for structural details.

\paragraph{Scope and Future Directions}
Nevertheless, the current formulation has limitations tied to its idealized assumptions. The closed-form travel-time prior is tailored for single-pipeline configurations in homogeneous or weakly heterogeneous backgrounds, which simplifies real-world subsurface clutter. To extend this framework to multi-target environments, the response-domain prior can be expanded by constructing multiple concurrent travel-time curves or a dense travel-time field. For complex stratified or laterally varying media, the closed-form hyperbolic approximation can be seamlessly replaced by ray-based tracking, or numerically estimated travel-time fields to update the response-domain channels. Transitioning from these synthetic, single-pipeline abstractions to stratified media, highly cluttered scenes, and ultimately, real-world GPR measurements remains a vital and promising direction for future work.

\section{Conclusion}

In this paper, we presented PCFlow, a physics-conditioned flow matching framework for fast and controllable GPR B-scan image synthesis. PCFlow constructs a Maxwell-informed dense condition field from parameterized physical simulation models by combining scene-domain physical cues with response-domain priors. This representation provides an interpretable mapping interface between subsurface scene parameters and radar response geometry. With SPADE-based physical modulation and classifier-free guidance (CFG), PCFlow guides conditional flow matching in the VAE latent space and strengthens condition-specific velocity components. This improves the alignment between generated radargrams and the encoded physical conditions.
Experiments on gprMax-based data demonstrate that PCFlow significantly improves response-geometry consistency without sacrificing visual fidelity, particularly in high-conductivity OOD scenarios. Ablation studies validate the necessity of response-domain priors for hyperbolic geometry preservation. Future directions include layered media, cluttered multi-target scenes, real GPR measurements, and adaptive guidance strategies.

\balance
\bibliographystyle{IEEEtran}
\bibliography{references}

\clearpage
\appendices

\section{Supplementary Derivations for the Maxwell-Informed Encoder}
\label{app:maxwell_encoder}

This appendix provides supporting derivations and implementation details for
the Maxwell-informed condition encoder in Sec.~\ref{sec:maxwell_encoder}. The main text already
defines the delayed-wavelet response model in \eqref{eq:bscan_master}, its
single-pipeline reduction in \eqref{eq:gpr_response_model}, the shift-based
travel-time error relation in \eqref{eq:curve_lower_bound}, and the approximate
pipeline two-way travel-time curve in \eqref{eq:twtt}. Here we only provide the
derivation path and the channel-level specification needed for reproducibility,
without repeating all implementation constants.

\subsection{Maxwell-Related Propagation Quantities}
\label{app:maxwell_quantities}

For the non-magnetic media considered in this work, we assume
$\mu\approx\mu_0$. The relative permittivity $\epsilon_r$ mainly controls wave
velocity and phase accumulation, while the conductivity $\sigma$ controls
propagation loss. In a locally homogeneous background, the effective
propagation velocity is
\begin{equation}
    v_{\mathrm{bg}}
    =
    \frac{1}{\sqrt{\mu_0\epsilon_0\epsilon_{r,\mathrm{bg}}}}
    =
    \frac{c_0}{\sqrt{\epsilon_{r,\mathrm{bg}}}} .
    \label{eq:app_vbg}
\end{equation}
Thus, a larger relative permittivity shifts the target response to a later
arrival time in the B-scan.

Conductivity introduces attenuation. Under the low-loss approximation for
$\epsilon=\epsilon_0\epsilon_r$, the attenuation coefficient is
\begin{equation}
    \alpha
    \approx
    \frac{\sigma}{2}
    \sqrt{\frac{\mu}{\epsilon}} .
    \label{eq:app_alpha}
\end{equation}
The wave impedance is approximately
\begin{equation}
    Z
    =
    \sqrt{\frac{\mu}{\epsilon}}
    \approx
    \frac{Z_0}{\sqrt{\epsilon_r}},
    \qquad
    \Gamma
    =
    \frac{Z_2-Z_1}{Z_2+Z_1}.
    \label{eq:app_impedance_reflection}
\end{equation}
These quantities motivate the material, velocity, attenuation, and phase,
impedance-edge, and response-strength channels used in the condition field.

\subsection{From Born Scattering to a Delayed-Wavelet B-Scan Model}
\label{app:born_master}

We start from a first-order Born scattering representation. In the frequency
domain, the scattered response between a transmitter at $\mathbf r_j$ and a
receiver at $\mathbf r_i$ can be written schematically as
\begin{equation}
    G_s^B(\mathbf r_i,\mathbf r_j,\omega)
    =
    \omega^2\mu
    \int_{\Omega}
    G_0(\mathbf r_i,\mathbf r,\omega)
    \delta\bar\epsilon(\mathbf r,\omega)
    G_0(\mathbf r,\mathbf r_j,\omega)
    d\mathbf r,
    \label{eq:app_born}
\end{equation}
where $G_0$ is the background Green function and
$\delta\bar\epsilon$ denotes the electromagnetic contrast.

Under a high-frequency approximation, the background Green function has the
form
\begin{equation}
    G_0(\mathbf r_2,\mathbf r_1,\omega)
    \approx
    A(\mathbf r_2,\mathbf r_1,\omega)
    \exp\!\left(i\omega\tau(\mathbf r_2,\mathbf r_1)\right),
    \label{eq:app_hf_green}
\end{equation}
where $A$ absorbs geometric spreading, attenuation, antenna response, and
polarization effects, and $\tau(\mathbf r_2,\mathbf r_1)$ is the travel time.
Substituting \eqref{eq:app_hf_green} into \eqref{eq:app_born} yields a kernel
whose phase is governed by the two-way travel time from the transmitter to
scatterer and from scatterer to receiver.

For a practical GPR trace indexed by acquisition position $a$, the source
spectrum, antenna response, amplitude factors, and polarization terms can be
collected into $H_a$. This gives the acquisition-wise frequency-domain form
\begin{equation}
    D_a(\omega)
    =
    \int_{\Omega}
    H_a(\mathbf r,\omega)
    \chi(\mathbf r)
    \exp\!\left(i\omega\tau_a(\mathbf r)\right)
    d\mathbf r,
    \label{eq:app_freq_master}
\end{equation}
where $\chi(\mathbf r)$ is the effective contrast and
$\tau_a(\mathbf r)$ is the corresponding two-way travel time.

Let $H_a(\mathbf r,\omega)=\Psi(\omega)W_a(\mathbf r,\omega)$, where
$\Psi(\omega)$ is the source wavelet spectrum. Under a narrowband or
weak-dispersion approximation, $W_a(\mathbf r,\omega)$ varies slowly over the
effective bandwidth and is approximated by $W_a(\mathbf r)$. Applying the
inverse Fourier transform gives
\begin{equation}
    d_a(t)
    =
    \int_{\Omega}
    W_a(\mathbf r)
    \chi(\mathbf r)
    \psi\!\left(t-\tau_a(\mathbf r)\right)
    d\mathbf r
    +
    \rho_a(t),
    \label{eq:app_time_master}
\end{equation}
where $\psi$ is the time-domain wavelet and $\rho_a$ denotes residual terms not
captured by the first-order approximation.

A B-scan arranges acquisition position along the horizontal coordinate $x$ and
time along the vertical coordinate $y$. Therefore, \eqref{eq:app_time_master}
leads to the delayed-wavelet B-scan model used in the main text in
\eqref{eq:bscan_master}. This model shows that response localization in a
radargram is controlled by both electromagnetic contrast and two-way travel
time.

\subsection{Single-Pipeline Response and Response-Domain Priors}
\label{app:single_pipeline_prior}

For the buried-pipeline setting, the earliest dominant response is associated
with a localized effective scattering region near the pipe. Approximating the
contrast by a localized scatterer reduces the delayed-wavelet integral to the
single-response model in \eqref{eq:gpr_response_model}, where $\tau_s(x)$ is
the dominant two-way travel-time curve, $A_s(x)$ is the effective response
amplitude, and $r_s$ collects residual full-wave effects.

The path lengths from the transmitter and receiver to the dominant scattering
center are
\begin{equation}
\begin{aligned}
    R_{\mathrm{tx}}(x)
    &=
    \sqrt{(x_{\mathrm{tx}}-x_c)^2+d^2},\\
    R_{\mathrm{rx}}(x)
    &=
    \sqrt{(x_{\mathrm{rx}}-x_c)^2+d^2}.
\end{aligned}
\label{eq:app_path_lengths}
\end{equation}
Using the effective background velocity in \eqref{eq:app_vbg}, this gives the
two-way travel-time prior in \eqref{eq:twtt}. For a monostatic or
near-monostatic configuration, this curve becomes the standard hyperbolic
response with apex arrival time $2d/v_{\mathrm{bg}}$.

The response amplitude is approximated by a compact path-length-dependent
attenuation and spreading term:
\begin{equation}
    A(x)
    \propto
    \frac{\exp[-\alpha_{\mathrm{bg}}L(x)]}
    {1+L(x)^2},
    \qquad
    L(x)=R_{\mathrm{tx}}(x)+R_{\mathrm{rx}}(x).
    \label{eq:app_amp_prior}
\end{equation}
This approximation motivates the response attenuation prior used in the
condition field. It provides only coarse amplitude guidance; waveform polarity,
interference, antenna effects, finite-size scattering, and other residual
full-wave details are learned by the conditional velocity network.

\section{Travel-Time Error Lower Bound}
\label{app:travel_time_lower_bound}

This section provides the derivation behind the lower-bound-style relation in
\eqref{eq:curve_lower_bound}. Ignoring the residual term, define
$I_0(y,x)=A_s(x)\psi(y-\tau_s(x))$ and
$\widehat I_0(y,x)=A_s(x)\psi(y-\widehat\tau_s(x))$. Their squared difference
can be written as
\begin{equation}
    \|\widehat I_0-I_0\|_2^2
    =
    \int_x
    A_s(x)^2
    \left\|
    \psi(\cdot-\Delta\tau(x))-\psi(\cdot)
    \right\|_2^2
    dx,
    \label{eq:app_shift_error_exact}
\end{equation}
where $\Delta\tau(x)=\widehat\tau_s(x)-\tau_s(x)$. This expression is followed by
the change of variable $u=y-\tau_s(x)$ for each fixed scan position $x$.

Assume $\psi\in H^1(\mathbb R)$ and $\|\psi'\|_2>0$. For a sufficiently small
shift $\Delta$, the first-order Taylor expansion in $L^2$ gives
\begin{equation}
\begin{aligned}
    \psi(u-\Delta) = {} & \psi(u)-\Delta\psi'(u)+R_2(u,\Delta), \\
    & \text{where } \|R_2(\cdot,\Delta)\|_2=o(|\Delta|).
\end{aligned}
    \label{eq:app_taylor_shift}
\end{equation}
Therefore,
\begin{equation}
    \|\psi(\cdot-\Delta)-\psi(\cdot)\|_2^2
    =
    \Delta^2\|\psi'\|_2^2+o(\Delta^2).
    \label{eq:app_wavelet_shift_expansion}
\end{equation}
It follows that there exist $\delta>0$ and $c_\psi>0$ such that for all
$|\Delta|\le\delta$,
\begin{equation}
    \|\psi(\cdot-\Delta)-\psi(\cdot)\|_2^2
    \ge
    c_\psi\Delta^2 .
    \label{eq:app_shift_identifiability}
\end{equation}
A valid local choice is $c_\psi=\frac{1}{2}\|\psi'\|_2^2$ after choosing
$\delta$ sufficiently small.

Applying \eqref{eq:app_shift_identifiability} to
\eqref{eq:app_shift_error_exact} yields
\begin{equation}
    \|\widehat I_0-I_0\|_2^2
    \ge
    c_\psi
    \int_x
    A_s(x)^2
    |\widehat\tau_s(x)-\tau_s(x)|^2
    dx.
    \label{eq:app_weighted_tau_error}
\end{equation}
If $A_s(x)\ge A_{\min}>0$ on the main response support $\mathcal X$, then
\begin{equation}
    \|\widehat I_0-I_0\|_2^2
    \ge
    A_{\min}^2 c_\psi
    \|\widehat\tau_s-\tau_s\|_{L^2(\mathcal X)}^2 .
    \label{eq:app_main_lower_bound}
\end{equation}

For the actual response $I_s=I_0+r_s$, the residual term weakens this relation
by an amount controlled by $\|r_s\|_2$. Thus, when the dominant shifted-wavelet
component is not overwhelmed by residual full-wave effects,
\begin{equation}
    \|\widehat I_0-I_s\|_2^2
    \gtrsim
    A_{\min}^2 c_\psi
    \|\widehat\tau_s-\tau_s\|_{L^2(\mathcal X)}^2
    -
    O(\|r_s\|_2^2).
    \label{eq:app_residual_lower_bound}
\end{equation}
This explains why a visually plausible B-scan can still be physically
inconsistent when its dominant response curve is misplaced.

\section{Ricker Wavelet Case}
\label{app:ricker}

The local shift-identifiability assumption in
\eqref{eq:app_shift_identifiability} is satisfied by Ricker-like GPR pulses.
For a standard Ricker wavelet,
\begin{equation}
    \psi(t)
    =
    \left(1-2\pi^2 f_0^2t^2\right)
    \exp(-\pi^2 f_0^2t^2),
    \label{eq:app_ricker}
\end{equation}
where $f_0$ is the center frequency. Let $a=\pi^2 f_0^2$. Then
$\psi(t)=(1-2at^2)\exp(-at^2)$, and
\begin{equation}
    \psi'(t)
    =
    2at(2at^2-3)\exp(-at^2).
    \label{eq:app_ricker_derivative}
\end{equation}
The derivative is nonzero and square-integrable. Its squared $L^2$ norm is
\begin{equation}
    \|\psi'\|_2^2
    =
    \int_{-\infty}^{\infty}
    4a^2t^2(2at^2-3)^2\exp(-2at^2)\,dt
    =
    \frac{15}{4\sqrt{2}}\sqrt{\pi}\sqrt{a}.
    \label{eq:app_ricker_derivative_norm}
\end{equation}
Since $\sqrt a=\pi f_0$, one valid local shift constant is
\begin{equation}
    c_\psi
    =
    \frac{1}{2}\|\psi'\|_2^2
    =
    \frac{15\pi}{8\sqrt{2}}\sqrt{\pi}\,f_0 .
    \label{eq:app_ricker_cpsi}
\end{equation}
This verifies that, within a local shift range around the main lobe, a temporal misalignment of a Ricker-like pulse induces a nonzero quadratic response error.

\section{Condition Channel Specification}
\label{app:channels}

This section summarizes the implementation-level specification of the Maxwell-informed condition field. All scalar physical fields are clipped to predefined ranges and linearly normalized to $[0,1]$.

The response-domain channels are constructed from the travel-time prior in
\eqref{eq:twtt}. The corresponding B-scan row coordinate is
\begin{equation}
\begin{aligned}
    y_\tau(x)
    &=
    \operatorname{clip}
    \left(
    \frac{\tau(x)}{T_{\mathrm{win}}}(H-1),
    0,
    H-1
    \right),
\end{aligned}
\label{eq:app_response_row}
\end{equation}
where $T_{\mathrm{win}}$ is the temporal window of the B-scan.

The 26 channels are summarized in Table~\ref{tab:app_channel_spec}. Smooth
dense fields are downsampled by average pooling, while sparse or localized
structures are downsampled by max pooling. This preserves both low-frequency
material trends and sharp response-domain anchors at the VAE latent resolution.

\begin{table*}[!t]
\centering
\caption{Specification of the 26-channel Maxwell-informed condition field.}
\label{tab:app_channel_spec}
\scriptsize
\setlength{\tabcolsep}{3.5pt}
\renewcommand{\arraystretch}{1.12}
\resizebox{\textwidth}{!}{
\begin{tabular}{
c
p{0.14\textwidth}
p{0.50\textwidth}
p{0.14\textwidth}
c
}
\toprule
Ch. & Name & Compact definition / role & Type & Pooling \\
\midrule
0 & Row coordinate
& Normalized B-scan row coordinate $y/(H-1)$.
& Coordinate & Avg. \\

1 & Scan coordinate
& Normalized antenna scan coordinate $x/(W-1)$.
& Coordinate & Avg. \\

\midrule
2 & Relative permittivity
& Clipped and normalized $\epsilon_r$ map.
& Material & Avg. \\

3 & Conductivity
& Clipped and normalized $\sigma$ map.
& Material & Avg. \\

4 & Velocity/impedance proxy
& Normalized $1/\sqrt{\epsilon_r}$, proportional to local wave velocity and related to impedance in non-magnetic media.
& Material & Avg. \\

5 & Phase constant
& Normalized propagation phase constant $\beta$ computed from $\epsilon_r$, $\sigma$, and the operating frequency.
& Material & Avg. \\

6 & Attenuation constant
& Normalized attenuation coefficient $\alpha$ computed from $\epsilon_r$, $\sigma$, and the operating frequency.
& Material & Avg. \\

\midrule
7 & Cumulative attenuation
& Depth-wise accumulated attenuation, approximating propagation loss along the vertical direction.
& Propagation & Avg. \\

8 & Phase sine
& Sine embedding of cumulative phase accumulation.
& Propagation & Avg. \\

9 & Phase cosine
& Cosine embedding of cumulative phase accumulation.
& Propagation & Avg. \\

10 & Material edge
& Sobel-gradient response of normalized material fields, highlighting material discontinuities.
& Propagation & Max \\

11 & Impedance edge
& Sobel-gradient response of the impedance/velocity proxy, highlighting reflection-related discontinuities.
& Propagation & Max \\

\midrule
12 & Pipe mask
& Binary support mask of the pipeline target in the physical scene domain.
& Geometry & Max \\

13 & Pipe boundary
& Soft ring-shaped boundary map around the pipe surface.
& Geometry & Max \\

14 & Signed distance
& Normalized signed-distance-like map relative to the pipe boundary.
& Geometry & Avg. \\

15 & Boundary normal-x
& Horizontal component of the pipe boundary normal weighted by the boundary ring.
& Geometry & Avg. \\

16 & Boundary normal-y
& Vertical component of the pipe boundary normal weighted by the boundary ring.
& Geometry & Avg. \\

17 & Pipe radius
& Constant map encoding normalized pipe radius.
& Geometry & Avg. \\

18 & Pipe lateral position
& Constant map encoding normalized pipe center position.
& Geometry & Avg. \\

19 & Burial depth
& Constant map encoding normalized burial depth.
& Geometry & Avg. \\

\midrule
20 & Hyperbola heatmap
& Gaussian heatmap centered at the approximate response curve $y_\tau(x)$.
& Response prior & Max \\

21 & Apex prior
& Gaussian anchor around the approximate hyperbola apex.
& Response prior & Max \\

22 & Travel-time curve
& Normalized response-row coordinate $y_\tau(x)/(H-1)$ broadcast along the row dimension.
& Response prior & Avg. \\

23 & Signed residual to curve
& Normalized signed row-wise offset from the approximate response curve.
& Response prior & Avg. \\

24 & Response attenuation
& Hyperbola-supported attenuation prior combining path-length attenuation and geometric spreading according to \eqref{eq:app_amp_prior}.
& Response prior & Max \\

25 & Response strength
& Response attenuation prior modulated by material and impedance contrast strength.
& Response prior & Max \\

\bottomrule
\end{tabular}
}
\end{table*}

\end{document}